# Adapting Segment Anything Model (SAM) through Prompt-based Learning for Enhanced Protein Identification in Cryo-EM Micrographs

Fei He[1,2,3], Zhivuan Yang[4,†], Mingyue Gao[4,†], Biplab Poudel[1,2,3,†], Newgin Sam Ebin Sam Dhas[1,2,3,†], Rajan Gyawali[1,2], Ashwin Dhakal[1,2], Jianlin Cheng[1,2] and Dong Xu[1,2,3*]

[1]Department of Electrical Engineering and Computer Science, [2]NextGen Precision Health, [3]Bond Life Sciences Center, University of Missouri, Columbia, MO, 65211, USA

[4]School of Information Science and Technology, Northeast Normal University, Changchun Jilin 130017, China

[†]: Equal contribution
* Corresponding author: Dong Xu, xudong@missouri.edu

**Abstract:**

Cryo-electron microscopy (cryo-EM) remains pivotal in structural biology, yet the task of protein particle picking, integral for 3D protein structure construction, is laden with manual inefficiencies. While recent AI tools such as Topaz and crYOLO are advancing the field, they do not fully address the challenges of cryo-EM images, including low contrast, complex shapes, and heterogeneous conformations. This study explored prompt-based learning to adapt the state-of-the-art image segmentation foundation model Segment Anything Model (SAM) for cryo-EM. This focus was driven by the desire to optimize model performance with a small number of labeled data without altering pre-trained parameters, aiming for a balance between adaptability and foundational knowledge retention. Through trials with three prompt-based learning strategies, namely head prompt, prefix prompt, and encoder prompt, we observed enhanced performance and reduced computational requirements compared to the fine-tuning approach. This work not only highlights the potential of prompting SAM in protein identification from cryo-EM micrographs but also suggests its broader promise in biomedical image segmentation and object detection.

# 1. Introduction

Protein particle identification from cryo-EM micrographs, a pivotal procedure for locating and extracting individual proteins from micrographs, enables the construction of 3D structures for downstream biological studies. Manual selection, however, is labor-intensive, time consuming, and susceptible to errors. Challenges are amplified due to the variable shapes, sizes, and orientations of proteins, compounded by prevalent ice contaminations, carbon edges, protein aggregates, and distorted proteins. Artificial Intelligence (AI)-driven image analysis has illuminated a pathway for protein picking from cryo-EM images, minimizing human intervention and mitigating bias and inconsistency. Recent strides in deep learning have underscored its potential in protein picking with AI approaches like crYOLO [1] and Topaz [2]. These methods conceptualized protein picking from cryo-EM micrographs as an image segmentation task, harnessing powerful deep neural networks for the challenge. However, the nuances of cryo-EM images, such as low contrast and diverse conformations, mean existing techniques, when trained on limited data, remain distant from precise protein identification.

Segment Anything Model (SAM) [3] has been released recently as a landmark advancement by Meta AI Research. Built upon an interactive prompt approach for diverse segmentation tasks and trained on the vast SA-1B dataset with over 11M images and 1B masks, SAM allows interactive prompts for user-driven object region proposals, surpassing previous image segmentation techniques, particularly in few-shot and zero-shot scenarios. However, a recent study [4] and our preliminary trials suggest that SAM, primarily trained on natural scene images, may not inherently cater to the intricacies and low contrast of cryo-EM images. Thus, devising tailored adaptation strategies to realize SAM's full potential for domain-specific applications remains an intriguing research avenue.

Fine-tuning is a predominant adaptation strategy to customize a pre-trained model for a specific task and domain. It adapts a model initially trained on a vast dataset and subsequently retraining it on a smaller, task-specific dataset. This process refines the model's ability to discern features and patterns relevant to the new task. The model's weights are updated according to the new dataset, yet much of the foundational knowledge remains intact. This strategy has garnered widespread adoption across fields like computer vision [5, 6] and natural language processing [7]. Several studies have opted to entirely or partially fine-tune SAM for domain-specific applications, such as medical image segmentation [8]. Given its size, fine-tuning SAM necessitates significant weight adjustments and expensive computational resources. Further, if the task-specific dataset is too small, there is a potential risk of overfitting.

A potential solution to this challenge is prompt-based tuning, which mandates no modifications to the pre-trained model. Instead, it incorporates additional tunable tensors or layers, enabling the model to adapt to new tasks. These tensors ensure task-specific refinement without modifying the original model's parameters, consequently diminishing overfitting risks. Rooted in NLP with outstanding performance in various scenarios, prompt-based learning made its foray into computer vision (CV) to assist specific applications. Visual Prompt Tuning (VPT) [9] represents a groundbreaking, parameter-efficient methodology. It pioneers the use of task-specific learnable prompts, demonstrating that one can obtain comparable performance to full fine-tuning by training only a small amount of model parameters. Building on the principles of VPT, researchers such as Wu et al. (2022) [10], and Wang et al. (2023) [11] extended the application of visual prompts to image generation tasks. In a parallel vein, Liu et al. (2023) [12], and Chen et al. (2023) [13] incorporated additional trainable parameters into the visual foundational models, enhancing their capabilities for a range of downstream tasks. Yet, no endeavors have been made to design adaptable prompts for SAM, particularly in niche scenarios like protein picking from cryo-EM micrographs.

While SAM's [3] interactive prompts have proven beneficial, there remains a demand for automated low-contrast image segmentation applications without manual interaction like cryo-EM protein picking. In our investigation, we have looked into prompt-based learning methodologies for SAM, centering on their utility for automated protein picking from cryo-EM micrographs. Our study has illuminated three prompt-based techniques: head prompt, prefix prompt, and encoder prompt, as Figure 1 described, all aimed at enabling SAM to pick proteins from cryo-EM micrographs without relying on interactive prompts. Each strategy engages SAM at a different stage but consistently showcases superior efficiency, requiring fewer computational resources and annotated samples than traditional fine-tuning. Our contributions not only offer an efficient avenue for cryo-EM protein picking but also furnish a comprehensive exploration of the advantages and versatility of prompt-based techniques within the biomedical imaging domain.

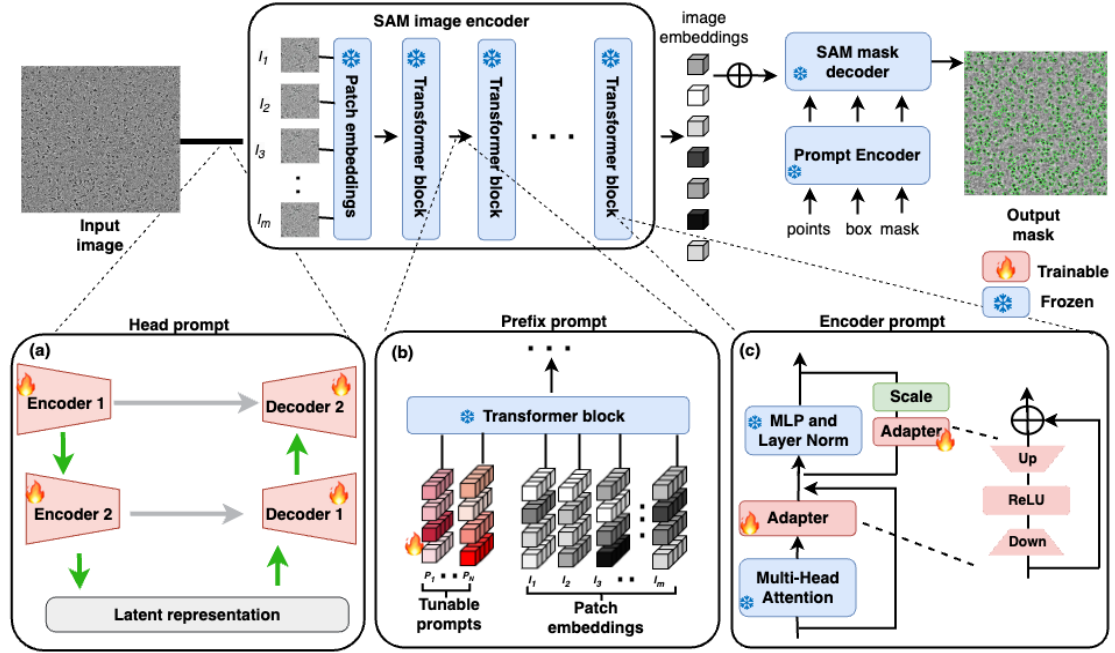

**Figure 1. Overview of the three proposed prompt-based learning strategies for SAM** (a) Head prompt: a trainable U-Net to enhance input micrographs for SAM adaptation. During training process, only the U-Net undergoes refinement, while the pretrained SAM is fixed. (b) Prefix Prompt: a set of tunable prompts that are prepended to image embeddings prior to their introduction into any transformer blocks in the image encoder. Training involves updating solely these modifiable prompts, with the entire SAM architecture retained in its original state. (c) Encoder Prompt: two adjustable autoencoder-like adapters integrated into any transformer blocks. Only the parameters of the adapters are updated during training while keeping SAM parameters frozen.

## 2. Method

### 2.1 Preliminary: SAM architecture

SAM integrates three pivotal components: an image encoder, a prompt encoder, and a mask decoder. The image encoder leverages the standard Vision Transformer (ViT) [14] to obtain image embeddings. Given SAM's ViT configuration with $N$ layers (in this study, we selected ViT-h taking $N = 32$ Transformer layers), an input image $I$ is partitioned into $m \times n$ uniform patches $\{I_{i,j} \in \mathbb{R}^{3 \times h \times w} \mid i,j \in \mathbb{N}, 1 \leq i \leq n, 1 \leq j \leq m\}$. $h, w$ signify the patch's height and width respectively. Each patch subsequently is embedded into a $d$-dimensional latent space, complemented by positional encoding:

$$e_0^{i,j} = Embed(I_{i,j}) \qquad e_0^{i,j} \in \mathbb{R}^d, i = 1,2 \ldots n, j = 1,2 \ldots m. \quad (1)$$

The image patch embeddings are denoted as $E_i = \{e_i^j \in \mathbb{R}^d \mid j \in \mathbb{N}, 1 \leq j \leq m\}$. These image embeddings serve as input to the $(k+1)$-th Transformer layer ($L_{k+1}$). The entire ViT can be formulated as:

$$E_k = L_k(E_{k-1}) \qquad k = 1,2, \ldots N \quad (2)$$

Each layer $L_k$ integrates Multi-head Self-Attention (MSA) [15], Multi-layer Perceptron (MLP), accompanied by LayerNorm and residual connections [14].

The prompt encoder translates user-defined regions of interest within images into prompt embeddings. User interactions, whether as points, boxes, or masks, are embedded based on their spatial coordinates. By default, SAM encompasses a grid covering all patches when no interaction prompts are provided. Finally, the mask decoder synergizes the outputs $E_I$ from the image encoder and $E_P$ from the prompt encoder to decode the final segmentation mask through $D$ two-way attention blocks [3], each containing two cross-attention layers $L_d$:

$$E_I^d, E_P^d = L_{d-1}(E_I^{d-1}, E_P^{d-1}) \quad (3)$$

$$E_P^{d+1}, E_I^{d+1} = L_d(E_P^d, E_I^d) \qquad d = 2, \ldots, D-1$$

Here the *(d-1)*-th cross-attention layers $L_{d-1}$ attends image embedding $E_I^{d-1}$ to prompt embedding $E_P^{d-1}$, and then *d*-th cross-attention layers $L_d$ prompt embedding $E_P^d$ attends to image embedding $E_I^d$. Through this iterative mapping, both types of embeddings are fused together to the image embedding $E_P^D$, which will be used to predict the final segmentation mask by an MLP as follows:

$$mask = MLP(E_P^D) \qquad (4)$$

**2.2 Head prompt**

The first prompt-based learning strategy involves constructing an autoencoder-like prompt to transform input cryo-EM micrographs into images optimized for SAM. To that end, we utilize a two-layer U-Net [16], which functions as a prompt on top of SAM, processing the input image *I* as follows:

$$\begin{aligned} f_1 &= \downarrow \sigma(I * Conv_1) \\ f_2 &= \downarrow \sigma(f_1 * Conv_2) \\ f_3 &= \sigma([\uparrow f_2 \cdot f_1] * Deconv_1) \\ I' &= \sigma([\uparrow f_3 \cdot I] * Deconv_2) \end{aligned} \qquad (5)$$

Here, Conv and Deconv describe the convolution layers and deconvolution layers at the encoder and decoder of U-net, respectively. $\sigma$ is RELU activation and f denotes the feature maps. $\uparrow$ and $\downarrow$ are up-sampling and down-sampling operations, respectively. [.] represents concatenation of two feature maps. The output image $I'$ is passed to SAM for image segmentation. During training, adjustments are made to the convolutional layers and deconvolution layers while keeping SAM unchanged. This implementation of U-Net, designed to enhance low contrast cryo-EM micrographs for improved SAM compatibility, is referred to as the 'head prompt'. It is worth noting that this head prompt is different from a recent work [4] of training a specialized attention-gated U-net on a cryo-EM micrograph dataset to predict protein particles first and then using its output as input for SAM to pick protein particles.

**2.3 Prefix prompt**

Drawing inspiration from Visual Prompt Tuning (VPT) [9], we introduced prompt tensors directly into the image embeddings preceding any Transformer layers within SAM's image encoder. By integrating a set of tunable prompt tensors $P_{ij} = \{p_{i,j}^k \in \mathbb{R}^d | i, j, k \in \mathbb{N}, 1 \leq k \leq N, 1 \leq i \leq n, 1 \leq j \leq m\}$ into the output from the $(k-1)$-th Transformer Layer $L_{k-1}$, Equation (2) evolves to:

$$[\_, E_k] = L_i([P_{k-1}, E_{k-1}]) \qquad k = 1, 2, \ldots N \qquad (6)$$

These tunable prompt tokens $P_{k-1}$ are prepended to the intermediate image embeddings $E_{k-1}$ from Transformer layer $L_{k-1}$ termed as prefix prompts. Throughout the training process, only these prompts $P_{ij}$ will be updated while the native SAM is frozen. Under the supervision of annotated images, the prompts $P_{ij}$ are optimized and they will provide the adaptability to the image embedding by the self-attention operation of the Transformer layer $L_k$, resulting in the adaptive image embedding $E_k$ for subsequent learning phases.

**2.4 Encoder prompt**

Building upon a strategy inspired by Wu et al. [17], we introduced two adapters within the Transformer layers, as illustrated in Figure 1c. These autoencoder-like adapters utilized a combination of an MLP and a RELU activation to condense *d*-dimensional image embeddings into a more compact *s*-dimensional format ($s \ll d$). Subsequently, another MLP assists in recovering this to an adaptive d-dimensional image embedding. As a result, Equation (2) is updated as:

$$\begin{aligned} E_k' &= Adapter_{k-1}(Atten_{k-1}(E_{k-1})) \\ E_k &= MLP_{k-1}(E_k') + \alpha \cdot Adapter_{k-1}'(E_k') \end{aligned} \qquad k = 1, 2, \ldots N \qquad (7)$$

Here $\alpha$ is a scale factor (we set $\alpha$=0.5 in our study by default). These two adapters will project the image embedding from native SAM to adaptive subspaces and then reconstruct an optimal image embedding at those subspaces. We named this the "encoder prompt" because it tailors image embeddings for specific tasks. During the training process, only the adapters will be updated as well.

**2.5 CryoPPP dataset and Metrics**

Cryo-EM micrographs are obtained by rapidly freezing of biological macromolecules in a thin layer of vitreous ice and imaging the frozen samples using the cryo-electron microscope to capture high-resolution images as shown in Figure 2. Firstly, biological samples such as purified proteins are prepared and embedded in a tiny layer of vitreous ice and loaded into a specialized grid. Then the grid is placed into a cryo-electron microscope that generates a focused electron beam to image the sample and allows capturing a series of 2D-micrographs from various orientations. The collected micrographs undergo processing like motion correction, denoising and contrast enhancement and finally, the protein particles are picked.

To validate our proposed prompt-based learning methods, we utilized the CryoPPP dataset [18], which encompasses a diverse set of 27 protein types. This comprehensive dataset boasts 6,893 high-resolution micrographs, each meticulously annotated with protein coordinates. On average, each protein type is represented by approximately 300 images. Firstly, the labels were cross-verified using 2D protein class validation methods. Subsequently, the annotations were further validated through 3D density map assessments in line with gold standard protocols.

To assess our experimental outcomes, we employed the Dice coefficient [19] as our performance metric. The Dice coefficient is a prevalent evaluation measure in image segmentation tasks, gauging the similarity between the predicted segmented mask and the ground-truth mask:

$$Dice = \frac{2*|segments \cap GT|}{|segments|+|GT|} \quad (8)$$

where '*segments*' refers to the pixels in the predicted mask, while '*GT*' represents pixels in the ground truth mask. A Dice score of 1 indicates a perfect alignment between the predicted and ground-truth masks, whereas a score of 0 denotes no overlap between them.

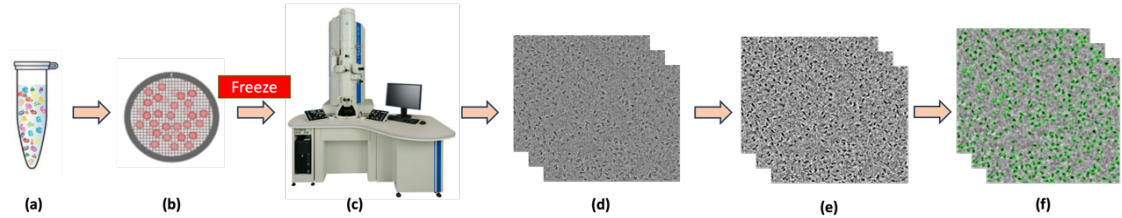

**Figure 2. The pipeline of cryo-EM micrograph collection and annotation.** (a) Preparing the protein sample and purification, (b) Applying a thin layer of a sample to an EM grid followed by removal of excess liquid and rapid plunge-freezing in liquid ethane to achieve cryogenic temperatures, (c) Calibration of the electron microscope to ensure accurate alignment, adjustment of the electron source's beam intensity, and configuration of camera exposure time and magnification, (d) Capture of raw images using the electron microscope, resulting in movie stacks composed of a series of 2D micrographs recorded chronologically. These stacks might exhibit motion artifacts, including drift and beam-induced motion. (e) Performing the motion correction to align the individual micrographs in the movie stack, (f) Manual identification and extraction of discrete particles from the micrographs.

**2.6 Training Setup and Hyperparameters**

For the optimization phase, we employed the Adam optimizer [20], initializing with a learning rate of $10^{-5}$. This optimizer was used for tuning the proposed prompt-based learning methods as well as for the fine-tuning of SAM. As our objective function, we adopted the widely accepted Dice loss [19], as represented by Equation (9):

$$DiceLoss = 1 - Dice \quad (9)$$

Here, the Dice score is computed as per Equation (8). As the training process progresses and the loss diminishes, the learning rate is adaptively decreased, minimizing the risk of bypassing the global optimum. We specified the maximum number of training epochs at 100. Among all epochs, the checkpoint exhibiting the minimal loss was retained as the fine-tuned model. When incorporating prompt-based learning, all SAM model parameters were frozen; only the additional tunable tensors were adjusted in response to the gradient of the loss function.

## 3. Results

### 3.1 Preliminary results of applying SAM in cryo-EM protein identification

To automate the protein-picking process from micrographs, we directly fed the cryo-EM micrographs into SAM [3] without any additional training. We utilized SAM's auto-segmentation mask generator to produce segmentation masks for a provisional assessment. Subsequently, Dice scores were computed based on a comparison between manual annotations and the predicted masks. The results from the native SAM are shown in Figure 3.

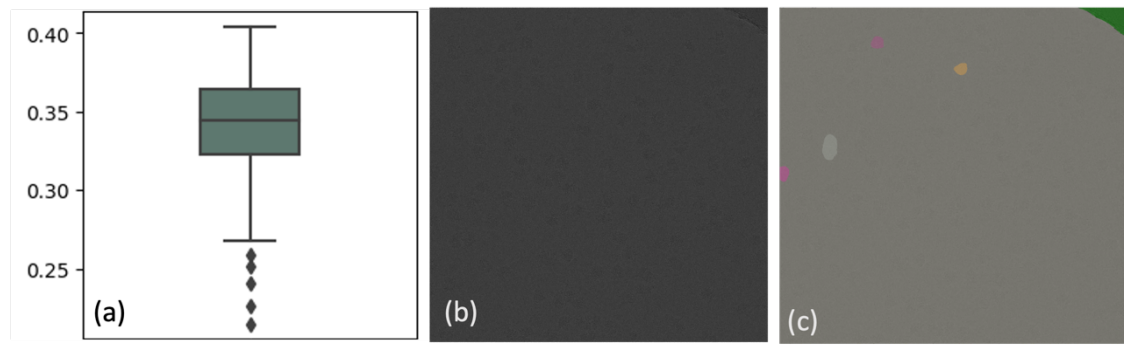

**Figure 3. Evaluation of native SAM's efficacy in protein identification using EMPIAR ID 10028 cryo-EM micrographs.** (a) Dice scores from predictions across all micrographs, (b) A representative example of a micrograph, (c) Corresponding segmentation mask generated by native SAM for the example micrograph.

Figure 3 presents the Dice scores and the segmentation masks generated by SAM's automatic mask generator for a sample protein type (EMPIAR ID: 10028). Rare particles were segmented successfully from the cryo-EM micrographs using native SAM. The Dice scores further indicated that the native configuration of SAM struggles to segment proteins from low-contrast images. A possible explanation is that SAM might have been trained predominantly on natural images rather than low-contrast ones. These results suggest the necessity of further adaption to unlock the potential of SAM in handling images with a low signal-to-noise ratio.

### 3.2 Prompt-based learning strategies robustly adapt SAM to cryo-EM protein identification

To assess the adaptability introduced by the three proposed prompt-based learning approaches to SAM, we conducted preliminary tests on three distinct protein types from the CryoPPP dataset [18] (EMPIAR IDs: 10028, 10947, and 10059). For the prefix prompt attempt, prefix tokens were inserted prior to each Transformer block within SAM's encoder. Concurrently, for the encoder prompt attempt, adapter blocks were incorporated into each Transformer block as well. To further probe the efficacy of these prompt-tuning strategies, we extracted varying sizes of training sets from these protein types, specifically 250, 200, 150, 100, 50, 30, 20, 10, and 5 micrographs. The remaining images were reserved for testing. This allowed us to evaluate the performance at diverse training costs. Figure 4 presents the Dice scores showcasing the results of the three prompt-based learning strategies across the different training sets for each of the three protein types.

Using the results from the protein EMPIAR ID 10028 as a representative example, the average Dice scores ascended to 0.823 with the inclusion of an adequate training set. This suggests that the three introduced prompt-based learning strategies effectively enable SAM to discern discriminative features from cryo-EM micrographs without changing any parameters of the native SAM. More interestingly, upon visualizing images post the head-prompt U-net process, as depicted in Figure 5, we noticed that the protein positions were highlighted with vivid contrasting colors in the prompted visualization. The head prompt approach facilitates the transformation of cryo-EM micrographs into more natural input images, thereby aligning SAM's ability to segment proteins from them. As the number of training samples decreased, both the head prompt and prefix prompt methods exhibited stable performance. However, when the training size was reduced to 5, encoder prompt crashed in the case of EMPIAR ID 10028. This might be attributable to the increased number of parameters in the encoder prompt approach, necessitating a minimum threshold of training sample size to function optimally. Consistent trends can be observed from other two types 10947 and 10059 of proteins.

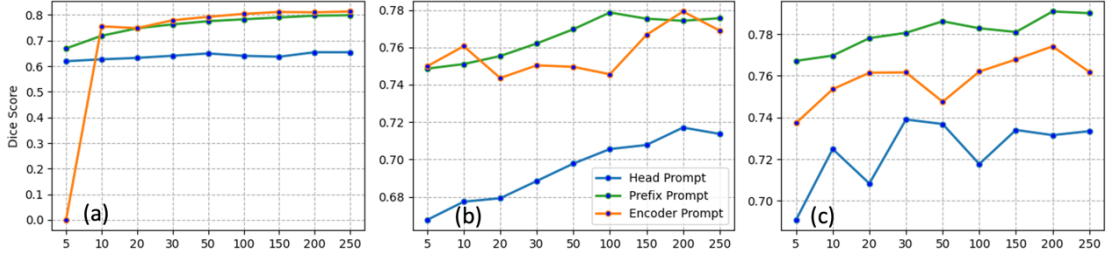

**Figure 4.** Dice score comparisons of SAM with head prompt, prefix prompt, and encoder prompt on different training sizes using proteins from (a) EMPIAR ID 10028 (b) EMPIAR ID 10947 (c) EMPIAR ID 10059

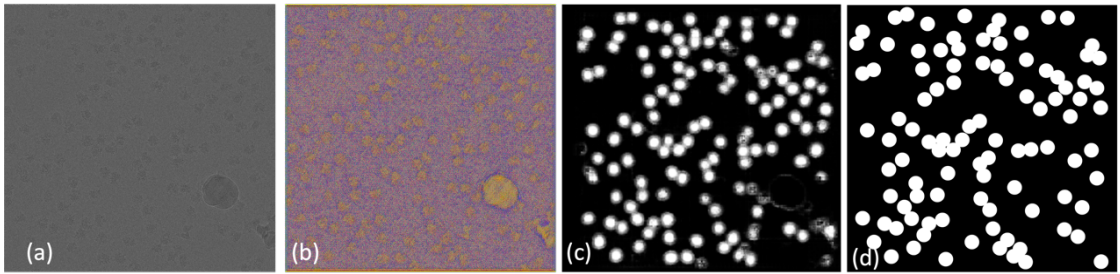

**Figure 5. A sample visualization of head prompt outputs** (a) input cryo-EM micrograph (b) prompted visualization after applying head prompt prior to SAM image encoder (c) segmentation mask from head prompted SAM (d) manual annotation to the input.

To assess the stability of our proposed prompt-based learning methods, we executed 10 separate evaluation tests on them, adhering to a uniform protocol. Initially, we segregated 200 images of protein type EMPIAR ID 10028 for testing to control the comparability of outcomes. Subsequently, from the remaining data, we randomly selected 10 images for model training using head prompt, prefix prompt, and encoder prompt techniques. The trained models were then evaluated using the testing data. This procedure was reiterated for 10 rounds. We analyzed the variation in Dice scores across these rounds for each test sample to measure the robustness of three proposed prompt-based learning methods. We provided an overview of the variance observed throughout the 10 rounds in Supplementary Figure 1. It reveals that none of the average variance of the three prompt-based methods exceeds 0.008, underlining their stable performance.

**3.3 Ablation test for prefix prompt and encoder prompt**

For the prefix and encoder prompt methods, a crucial hyper-parameter to consider is the 'prompt depth', which determines the number of Transformer blocks in SAM's image encoder where prompts are inserted. Our utilized ViT in SAM has 32 Transformer blocks. We conducted ablation tests ranging from inserting prompts into all Transformer blocks to including them in some topmost or bottommost Transformer blocks. We present a graph illustrating the average Dice scores across these variations in Supplementary Figure 2. In this figure, the prefix prompt displayed consistent performance regardless of prompt depth, reflecting SAM's inherent adaptability. The trade-off between performance and parameter usage was achieved by inserting prompt tokens into the bottommost (latter) block. This suggests that the latter blocks, closer to the output layer, receive stronger gradient signals, facilitating easier prompt tuning. Conversely, the performance of the encoder prompt method improved with increased prompt depth at the bottommost blocks. This indicates that a more extensive inclusion of encoder prompts makes SAM more promotive. Consequently, the greater the number of encoder prompts incorporated, the larger training size required to maximize its capabilities. Therefore, when implementing this strategy, it's crucial to match the prompt depth to the available training dataset. To strike a balance between computational efficiency and performance, we chose to insert both prefix and encoder prompts into the last Transformer block, deeming it the optimal depth for subsequent experiments.

**3.4 Comparison of finetuning and prompting approaches across all protein types**

To thoroughly assess the performance of the three proposed prompt-based learning strategies across all protein types from the CryoPPP dataset, we benchmarked them against the conventional fine-tuning method. For each protein type sourced from CryoPPP dataset we randomly selected 10 samples as the training set, with the remaining samples designated as the test set for SAM adaptation. Figure 6 showcases the average Dice scores, accompanied by error bars, derived from both the fine-tuning and the three prompt-based learning techniques. It illustrates the comparative performances of four adaptation techniques. Notably, the encoder prompt method outperformed across 12 protein types, demonstrating its efficacy when provided with sufficient training samples. Following closely, the prefix prompt method excelled in 7 types. The head prompt outperformed in 6 protein types, and it offered more interpretable prompts, also holding a potential precursor for other adaptation techniques.

Broadly speaking, prompt-based learning surpasses the fine-tuning approach for an impressive 92.6% (25 out of 27) of protein types from the CryoPPP dataset. In contrast, the fine-tuning method outperformed merely 2 protein types (10061 and 10289), but its superiority was minor compared to prompt-based learning strategies. We further examined the annotations of these 2 protein types and found relatively more false proteins caused by artifacts on their micrographs. Since SAM was incapable of categorizing segmented objects, such artifacts may cause misrecognitions by prompted SAM due to its sensitivity to particle detection from cryo-EM micrographs.

Table 1 presents the trainable parameters and GPU memory usage for each adaptation method. The results indicate that fine-tuning SAM is resource-intensive, requiring considerable computational power to adjust a vast number of parameters. Moreover, while fine-tuning SAM demands high-end GPUs such as A100, it struggles to yield impressive results with limited training data, as evidenced by prior experiments. On the other hand, the encoder prompt method, despite being more computationally efficient than fine-tuning, still acts with superior performance across most protein types, given adequate training data. Both the head prompt and prefix prompt methods introduce a few tunable parameters to SAM and thus are memory efficient. For these methods, mid-range GPUs like the GTX 1080Ti are adequate to manage. However, for those seeking enhanced performance via the encoder prompt method on SAM, more robust GPUs, such as the GTX 3090Ti or GTX 4090Ti, would be apt.

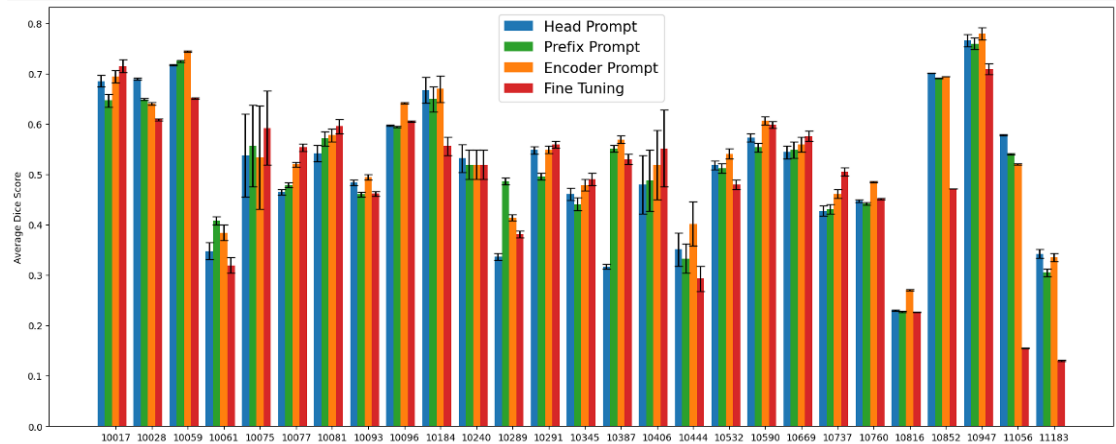

**Figure 6. Average Dice scores of SAM with head prompt, prefix prompt, encoder prompt, and fine-tuning across all protein types**

**Table 1 The number of trainable parameters and GPU memory usage for four different methods**

| Methods | Fine-tuning | Head prompt | Prefix prompt | Encoder prompt |
| --- | --- | --- | --- | --- |
| Trainable parameters | 4,058,340 | 410,019 | 2,621,440 | 52,531,200 |
| GPU memory | ~30G | ~12G | <16G | >40G |

## 3.5 Comparisons with existing tools

To further assess the effectiveness and generalizability of our proposed prompting methods, we compared our techniques with two accessible cryo-EM protein identification tools: crYOLO [1] and Topaz [2]. Initially, we curated an independent testing set comprising seven unique protein types from the CryoPPP dataset, leaving all remaining protein types for training. This partition was meticulously designed, accounting for variables such as protein classification, shape, size, and cumulative structural weight. The selected proteins spanned various categories, including transport, membrane, signaling, viral proteins, ribosomes, and aldolase, among others. These proteins exhibited diverse shapes—ranging from rod to circular—and their structural weights varied from 77 kDa to 2198 kDa. Supplementary Table 1 outlines the specifics of the training datasets, while Supplementary Table 2 details the independent testing dataset. Following this data partition, we employed the SAM model with our adaptation techniques (including finetuning, head prompt, prefix prompt, and encoder prompt) on the training protein types. When training with the adaptation techniques, we randomly selected specific subsets. This allowed us to compare the performance of the traditional tools, when trained on a full dataset, against our proposed methods trained on a partial dataset. Supplementary Table 3 illustrates the Dice scores for crYOLO, Topaz, and SAM equipped with finetuning, head prompt, prefix prompt, and encoder prompt on all independent testing protein types. The metrics for crYOLO and Topaz were determined using their respective default thresholds.

Table 2 displays the average Dice scores for Topaz, crYOLO, and SAM when utilizing head prompt, prefix prompt, encoder prompt, and finetuning on representative testing protein types of EMPIAR ID 10028, 10093 and 11056, at different training sizes. Notably, these scores surpass or are close to those observed for Topaz and crYOLO. This pronounced superiority depicts SAM's enhanced capability to detect proteins in low-contrast cryo-EM micrographs, even with a reduced training dataset size. Furthermore, the results confirm that all the adaptation strategies maintain SAM's generalization ability in identifying proteins from independent test samples. Among the various adaptations, the three proposed prompt-based learning methods mostly outperformed the finetuning approach across different training sizes, aligning with earlier experimental observations.

**Table 2. Average Dice scores for SAM with head prompt, prefix prompt, encoder prompt, finetuning, and existing methods (Topaz and CrYOLO) across independent testing protein types of EMPIAR ID 10028, 10093 and 11056**

| Testing EMPIAR ID | Training size | Head prompt | Prefix Prompt | Encoder Prompt | Finetuning | Topaz [2] | CrYOLO [1] |
|---|---|---|---|---|---|---|---|
| 10028 | 1000 | 0.616 | 0.714 | 0.698 | 0.702 | 0.616* | 0.356* |
|  | 1500 | 0.636 | 0.723 | 0.738 | 0.673 |  |  |
|  | 2000 | 0.642 | 0.726 | **0.752** | 0.712 |  |  |
|  | 4000 | 0.641 | 0.704 | 0.747 | 0.727 |  |  |
| 10093 | 1000 | 0.464 | 0.489 | 0.482 | 0.485 | 0.504* | 0.086* |
|  | 1500 | 0.467 | 0.511 | 0.498 | 0.490 |  |  |
|  | 2000 | 0.474 | 0.506 | **0.529** | 0.496 |  |  |
|  | 4000 | 0.460 | 0.506 | 0.511 | 0.491 |  |  |
| 11056 | 1000 | 0.464 | 0.434 | 0.464 | 0.338 | **0.692*** | 0.284* |
|  | 1500 | 0.518 | 0.445 | 0.481 | 0.422 |  |  |
|  | 2000 | 0.569 | 0.522 | 0.559 | 0.377 |  |  |
|  | 4000 | 0.572 | 0.518 | 0.560 | 0.430 |  |  |

*These results were obtained from the tools' inference on the same independent test data, without any additional training

We also examined the predictions from the three methods on specific micrographs to discern their distinctive characteristics. Figure 7 showcases the variations in particle picking among crYOLO [1], Topaz [2], and our proposed methods, using an individual cryo-EM micrograph representing each of two protein types (EMPIAR IDs 10345 and 11056). CrYOLO typically detects fewer proteins, whereas Topaz often over-sensitively identifies artifacts as proteins. While SAM with finetuning produces segmentation masks that more accurately pinpoint protein locations, there are occasional misidentifications. In contrast, SAM with head prompt, prefix prompt, and encoder prompt demonstrates a heightened precision, clarity, and robustness in protein localization.

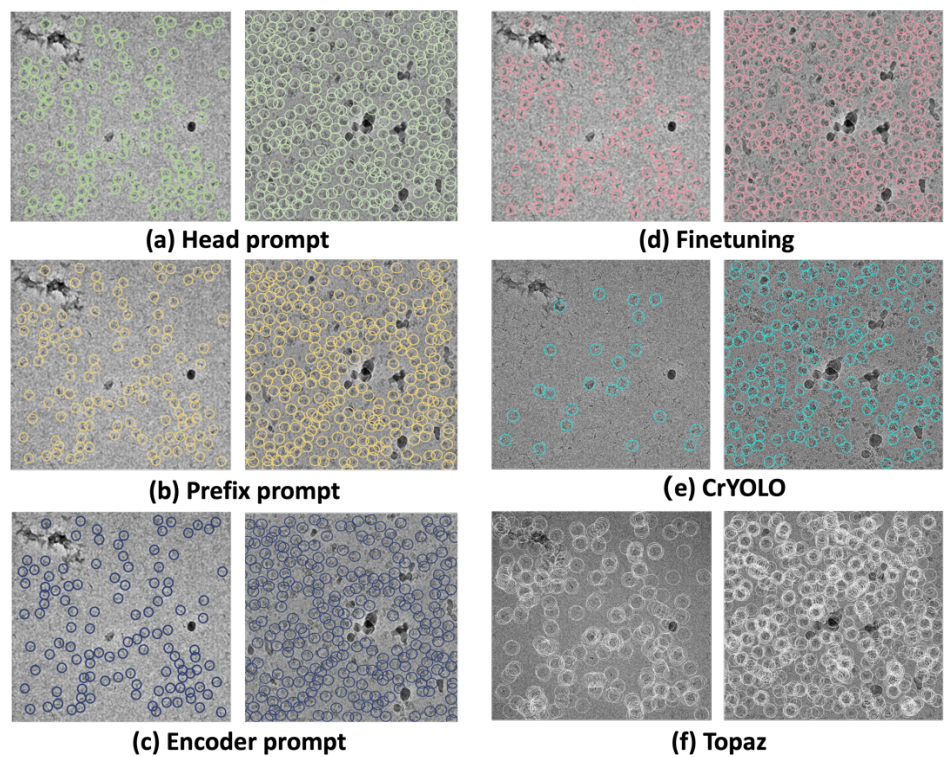

**Figure 7.** Visualization of protein particle identification on three sample micrographs achieved by SAM with (a) Head Prompt, (b) Prefix Prompt, (c) Encoder Prompt, (d) Finetuning, along with (e) CrYOLO, and (f) Topaz.

**4. Conclusion**

In this study, we introduced three innovative prompt-based learning strategies: prefix prompt, encoder prompt, and head prompt for detecting proteins within the CryoPPP dataset. These prompts were designed to bolster the adaptability of the Segment Anything Model (SAM). The tunable nature of the proposed prompt-based methods allows SAM to capture discriminative features without changing the core architecture of the native model. Furthermore, these strategies empower SAM to receive effective supervision from a limited set of training samples, steering the model to concentrate on the salient features essential for image segmentation. The comparison against conventional fine-tuning methods and existing tools, such as crYOLO and Topaz, reveals the strengths of the prompt-based learning methods in protein detection efficacy, requiring fewer computational resources and less labeled data. However, it is worth noting that SAM, in its original design, is not equipped to categorize objects post-segmentation, and there is a possibility for SAM to misinterpret artifacts as proteins within cryo-EM micrographs. As we move forward, our future investigations will delve into the feasibility of integrating a classifier to SAM to distinguish between genuine proteins and false particles, and we will also assess our proposed approaches' performance and versatility across various datasets. This research accentuates the transformative potential of prompt-based approaches in revolutionizing cryo-EM protein identification, setting the stage for broader applications in biomedical image segmentation and object detection.

**Acknowledgements**

We would like to acknowledge the valuable contribution from OpenAI's GPT-4 model for aiding in language editing. This work was funded by the National Institutes of Health [R35-GM126985, R01GM146340].

**Availability of Code and Data**

Source code is publicly available at https://github.com/yangyang-69/Prompt_sam_cryoPPP.git. CryoPPP dataset is also available at https://github.com/BioinfoMachineLearning/cryoppp.

## Appendix

There is one additional file containing Supplementary Figures 1-2 and Supplementary Tables 1-3.

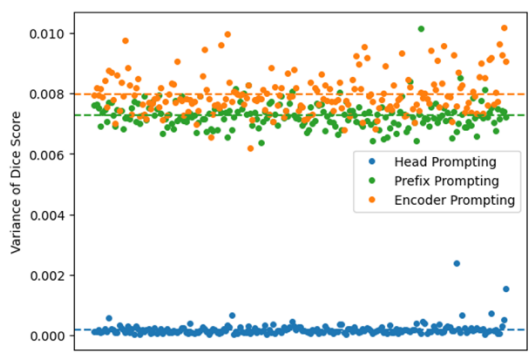

**Supplementary Figure 1. Variation of Dice scores of 200 testing samples from EMPIAR ID 10028 protein across 10-round validation for** head prompt, prefix prompt, and encoder prompt. Each dot describes the Dice score variance from a particular sample resulting from SAM with a proposed prompt-based learning approach.

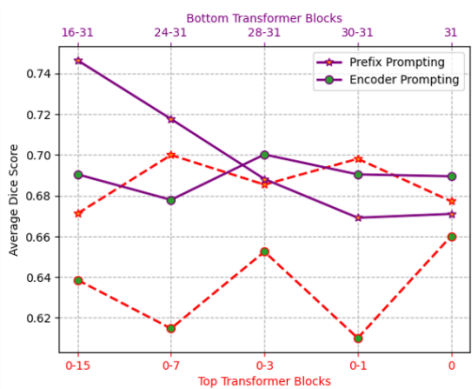

**Supplementary Figure 2. Comparative analysis of average Dice scores upon the integration of prefix prompts and encoder prompts at varied transformer layers within SAM.** Transformer blocks from 0 to 15 are denoted as top transformer blocks, proximate to the input layer, while blocks from 16 to 31 are termed as bottom transformer blocks, adjacent to the output layer. The red dotted line delineates the average Dice scores achieved by both prompt approaches across diverse combinations of top transformer blocks. Conversely, the purple solid line depicts the average Dice scores for various combinations involving the bottom transformer blocks.

**Supplementary Table 1. Meta-information of the training dataset for generalization test**

| EMPIAR ID | Protein Type | Image Size | Total Structure Weight (kDa) | Total images |
|---|---|---|---|---|
| 10005 | TRPV1 Transport Protein | (3710, 3710) | 272.97 | 29 |
| 10059 | TRPV1 Transport Protein | (3838, 3710) | 317.88 | 291 |
| 10075 | Bacteriophage MS2 | (4096, 4096) | 1000* | 299 |
| 10077 | Ribosome (70S) | (4096, 4096) | 2198.78 | 300 |
| 10096 | Viral Protein | (3838, 3710) | 150* | 300 |
| 10184 | Aldolase | (3838, 3710) | 150* | 296 |
| 10240 | Lipid Transport Protein | (3838, 3710) | 171072 | 299 |
| 10289 | Transport Protein | (3710, 3838) | 361.39 | 300 |
| 10291 | Transport Protein | (3710, 3838) | 361.39 | 300 |
| 10387 | Viral Protein | (3710, 3838) | 185.87 | 299 |
| 10406 | Ribosome (70S) | (3838, 3710) | 632.89 | 139 |
| 10444 | Membrane Protein | (5760, 4092) | 295.89 | 296 |
| 10526 | Ribosome (50S) | (7676, 7420) | 1085.81 | 220 |
| 10590 | TRPV1 Transport Protein | (3710, 3838) | 1000* | 296 |
| 10737 | Membrane Protein | (5760, 4092) | 155.83 | 292 |
| 10760 | Membrane Protein | (3838, 3710) | 321.69 | 300 |
| 10816 | Transport Protein | (7676, 7420) | 166.62 | 300 |
| 10852 | Signaling Protein | (5760, 4092) | 157.81 | 343 |
| 11051 | Transcription/DNA/RNA | (3838, 3710) | 357.31 | 300 |
| 11057 | Hydrolase | (5760, 4092) | 149.43 | 295 |
| 11183 | Signaling Protein | (5760, 4092) | 139.36 | 300 |

* Theoretical weights of the proteins.

**Supplementary Table 2. Meta-information of the independent testing dataset for generalization test**

| EMPIAR ID | Protein Type | Image Size | Total Structure Weight (kDa) | Total images |
|---|---|---|---|---|
| 10028 | Ribosome (80S) | (4096, 4096) | 2135.89 | 300 |
| 10081 | Transport Protein | (3710, 3838) | 298.57 | 300 |
| 10345 | Signaling Protein | (3838, 3710) | 244.68 | 295 |
| 11056 | Transport Protein | (5760, 4092) | 88.94 | 305 |
| 10532 | Viral Protein | (4096, 4096) | 191.76 | 300 |
| 10093 | Membrane Protein | (3838, 3710) | 779.4 | 298 |
| 10017 | β-galactosidase | (4096, 4096) | 450* | 84 |

* Theoretical weights of the proteins.

**Supplementary Table 3. Average Dice scores for SAM with head prompt, prefix prompt, encoder prompt, finetuning, and existing methods (Topaz and CrYOLO) on independent testing dataset**

| Testing EMPIAR ID | Training size | Head prompt | Prefix Prompt | Encoder Prompt | Fine Tuning | Topaz [2] | CrYOLO [1] |
|---|---|---|---|---|---|---|---|
| 10028 | 1000 | 0.616 | 0.714 | 0.698 | 0.702 | 0.616* | 0.356* |
|  | 1500 | 0.636 | 0.723 | 0.738 | 0.673 |  |  |
|  | 2000 | 0.642 | 0.726 | 0.752 | 0.712 |  |  |
|  | 4000 | 0.641 | 0.704 | 0.747 | 0.727 |  |  |
| 10093 | 1000 | 0.464 | 0.489 | 0.482 | 0.485 | 0.504* | 0.086* |
|  | 1500 | 0.467 | 0.511 | 0.498 | 0.490 |  |  |
|  | 2000 | 0.474 | 0.506 | 0.529 | 0.496 |  |  |
|  | 4000 | 0.460 | 0.506 | 0.511 | 0.491 |  |  |
| 11056 | 1000 | 0.464 | 0.434 | 0.464 | 0.338 | 0.692* | 0.284* |
|  | 1500 | 0.518 | 0.445 | 0.481 | 0.422 |  |  |
|  | 2000 | 0.569 | 0.522 | 0.559 | 0.377 |  |  |
|  | 4000 | 0.572 | 0.518 | 0.560 | 0.430 |  |  |
| 10081 | 1000 | 0.662 | 0.709 | 0.504 | 0.613 | 0.825* | 0.214* |
|  | 1500 | 0.684 | 0.712 | 0.636 | 0.587 |  |  |
|  | 2000 | 0.695 | 0.720 | 0.718 | 0.617 |  |  |
|  | 4000 | 0.713 | 0.708 | 0.677 | 0.636 |  |  |
| 10017 | 1000 | 0.518 | 0.582 | 0.496 | 0.528 | 0.694* | 0.041* |
|  | 1500 | 0.505 | 0.632 | 0.287 | 0.554 |  |  |
|  | 2000 | 0.606 | 0.653 | 0.647 | 0.514 |  |  |
|  | 4000 | 0.588 | 0.620 | 0.351 | 0.588 |  |  |
| 10345 | 1000 | 0.509 | 0.548 | 0.319 | 0.344 | 0.659* | 0.111* |
|  | 1500 | 0.526 | 0.561 | 0.396 | 0.352 |  |  |
|  | 2000 | 0.530 | 0.584 | 0.578 | 0.410 |  |  |
|  | 4000 | 0.565 | 0.589 | 0.494 | 0.439 |  |  |
| 10532 | 1000 | 0.279 | 0.538 | 0.299 | 0.333 | 0.757* | 0.239* |
|  | 1500 | 0.346 | 0.490 | 0.317 | 0.343 |  |  |
|  | 2000 | 0.281 | 0.554 | 0.432 | 0.309 |  |  |
|  | 4000 | 0.336 | 0.518 | 0.548 | 0.370 |  |  |

*These results were obtained from the tool's inference on the same independent test data, without any additional training